\documentclass{article}

\usepackage{microtype}
\usepackage{graphicx}
\usepackage{subcaption}
\usepackage{booktabs} 

\usepackage{hyperref}

\usepackage[preprint]{icml2026}

\usepackage{amsmath}
\usepackage{amssymb}
\usepackage{mathtools}
\usepackage{amsthm}

\usepackage[capitalize,noabbrev]{cleveref}

\usepackage{xcolor}
\usepackage{multirow}
\usepackage{graphicx}
\usepackage{booktabs}
\usepackage{tabularx}
\usepackage[flushleft]{threeparttable}
\usepackage{array}
\usepackage{pifont} 
\usepackage{booktabs}
\usepackage{textcomp}
\usepackage{makecell}
\usepackage{caption}
\usepackage{hyperref}
\usepackage{bm}
\usepackage[table]{xcolor}
\definecolor{mygreen}{HTML}{77a461}
\definecolor{myblue}{HTML}{6c8ebf}
\definecolor{myred}{HTML}{b85450}
\newcommand{\cmark}{\ding{51}}
\newcommand{\xmark}{\text{-}}

\theoremstyle{plain}

\theoremstyle{definition}

\theoremstyle{remark}

\usepackage[textsize=tiny]{todonotes}

\icmltitlerunning{DiffVC-RT: Towards Practical Real-Time Diffusion-based Perceptual Neural Video Compression}

\begin{document}

\twocolumn[
  \icmltitle{DiffVC-RT: Towards Practical Real-Time\\Diffusion-based Perceptual Neural Video Compression}



  \icmlsetsymbol{equal}{*}

  \begin{icmlauthorlist}
    \icmlauthor{Wenzhuo Ma}{whu}
    \icmlauthor{Zhenzhong Chen}{whu}
  \end{icmlauthorlist}

  \icmlaffiliation{whu}{School of Remote Sensing and Information Engineering, Wuhan University, Wuhan, China}

  \icmlcorrespondingauthor{Zhenzhong Chen}{zzchen@whu.edu.cn}

  \icmlkeywords{Neural Video Compression, Diffusion Model, Real Time}

  \vskip 0.3in
]



\printAffiliationsAndNotice{}  

\begin{abstract}
  The practical deployment of diffusion-based Neural Video Compression (NVC) faces critical challenges, including severe information loss, prohibitive inference latency, and poor temporal consistency. To bridge this gap, we propose \textbf{DiffVC-RT}, the first framework designed to achieve real-time diffusion-based perceptual NVC. First, we introduce an \textbf{\textit{Efficient and Informative Model Architecture}}. Through strategic module replacements and pruning, this architecture significantly reduces computational complexity while mitigating structural information loss. Second, to address generative flickering artifacts, we propose \textbf{\textit{Explicit and Implicit Consistency Modeling}}. We enhance temporal consistency by explicitly incorporating a zero-cost Online Temporal Shift Module within the U-Net, complemented by hybrid implicit consistency constraints. Finally, we present an \textbf{\textit{Asynchronous and Parallel Decoding Pipeline}} incorporating Mixed Half Precision, which enables asynchronous latent decoding and parallel frame reconstruction via a Batch-dimension Temporal Shift design. Experiments show that DiffVC-RT achieves \textbf{80.1\%} bitrate savings in terms of LPIPS over VTM-17.0 on HEVC dataset with real-time encoding and decoding speeds of \textbf{206 / 30 fps} for 720p videos on an NVIDIA H800 GPU, marking a significant milestone in diffusion-based video compression.
\end{abstract}

\begin{figure}[t]
  \vskip 0.2in
  \centering
  \centerline{\includegraphics[width=\columnwidth]{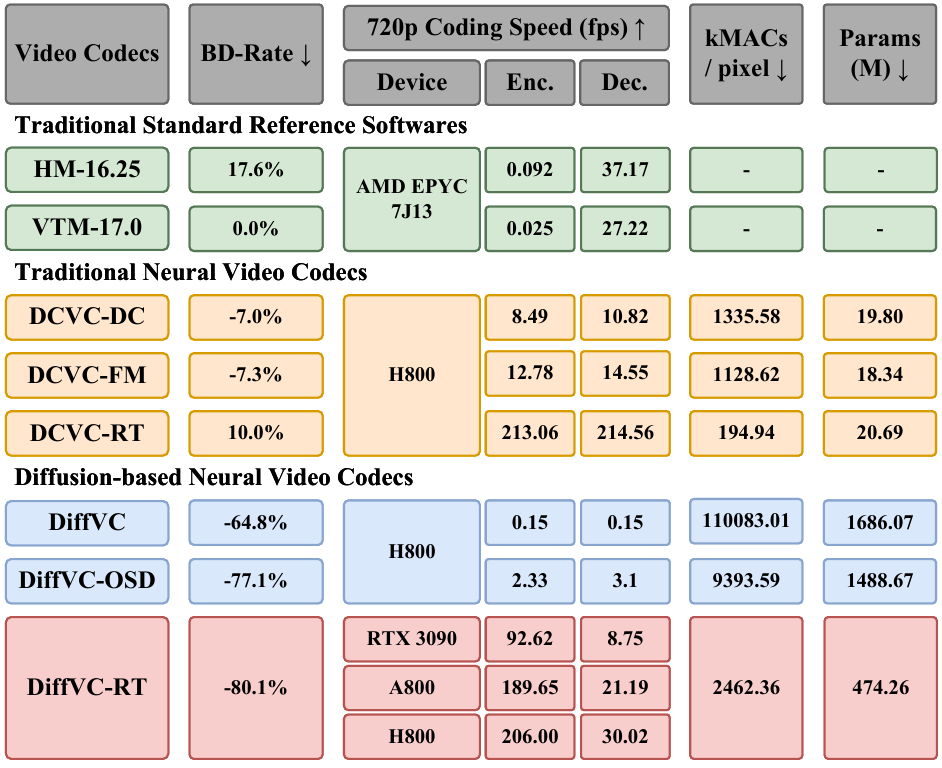}}
  \caption{
    Performance and efficiency overview of our proposed DiffVC-RT compared with various video codecs. The BD-Rate is measured using LPIPS on HEVC datasets, with VTM-17.0 serving as the anchor.
  }
  \label{fig:overview}
\end{figure}

\section{Introduction}
Driven by the rapid development of deep neural networks, Neural Video Compression (NVC) has demonstrated substantial potential to surpass conventional standards, such as H.265/HEVC~\cite{HEVC} and H.266/VVC~\cite{VVC}. Notably, DCVC-RT~\cite{DCVC-RT}, the latest iteration of the DCVC family~\cite{DCVC,DCVC-TCM,DCVC-HEM,DCVC-DC,DCVC-FM,DCVC-RT}, has achieved real-time encoding and decoding speeds while surpassing the VVC standard's performance, signaling that NVCs are steadily advancing towards practical application. However, the majority of existing NVC methods predominantly optimize for pixel-level distortion metrics (e.g., MSE). This objective often leads to visually unpleasant blurring and the loss of high-frequency details, particularly at low bitrates. To pursue superior subjective visual experience, perceptual NVC based on generative models has emerged as a significant research focus. In particular, foundation Diffusion Models~\cite{DDPM,VDM}, pre-trained on massive high-quality image-text datasets, leverage powerful generative priors to achieve remarkable texture reconstruction, as demonstrated in pioneering works like DiffVC~\cite{DiffVC} and DiffVC-OSD~\cite{DiffVC-OSD}. These methods have established a new paradigm for perception-oriented video compression.

Despite the promise of diffusion-based NVC, its deployment in real-world scenarios is hindered by serveral challenges:
\begin{itemize}
    \item \textbf{Severe Information Loss:} Existing diffusion-based methods typically inherit the VAE from Stable Diffusion~\cite{SD}, which downsamples frames by a factor of 8 and compresses them into only 4 channels. This narrow information bottleneck inevitably discards structural details.

    \item \textbf{Prohibitive Inference Latency:} Although some approaches attempt to accelerate inference via One-Step Diffusion, the substantial computational cost of the large-scale U-Net and VAE, combined with a serial, frame-by-frame processing paradigm, creates a significant barrier to real-time speed (e.g., 30 fps).
    
    \item \textbf{Poor Temporal Consistency:} Owing to the stochastic generative process of diffusion models, generating frames largely independently can cause temporal jitter and flicker, especially in textured regions.
\end{itemize}

To address the these challenges and bridge the gap between high perceptual quality and real-time throughput, we propose \textbf{DiffVC-RT}, the first framework designed for real-time diffusion-based perceptual neural video compression. We redesign the system across three dimensions: model architecture, consistency modeling, and the inference pipeline.

\begin{itemize}
    \item \textbf{Efficient and Informative Model Architecture:} We remove the complex motion branch and replace the compute-heavy VAE encoder with an efficient PixelUnshuffle operation to achieve real-time encoding. This is coupled with a Latent Channel Expansion strategy to maximize the preservation of structural information. Furthermore, by incorporating a pruned, lightweight U-Net and VAE Decoder, we significantly reduce the computational burden on the decoding side.
    
    \item \textbf{Explicit and Implicit Consistency Modeling:} By embedding an Online Temporal Shift Module into the residual blocks of the U-Net, we facilitate inter-frame interaction by shifting feature channels along the temporal dimension. This ``zero-parameter, zero-computation'' explicit temporal enhancement, complemented by hybrid implicit consistency constraints, effectively enhances the temporal stability.
    
    \item \textbf{Asynchronous and Parallel Decoding Pipeline:} Exploiting the decoupling between latent decoding and frame reconstruction, we devise an asynchronous execution strategy. We introduce a Batch-dimension Temporal Shift design to enable parallel frame reconstruction. Additionally, we adopt a Mixed Half Precision strategy to accelerate inference while maintaining numerical stability. Together, these designs bridge the final gap towards real-time decoding.
\end{itemize}

Extensive experiments across multiple benchmarks demonstrate that DiffVC-RT breaks the speed bottleneck of diffusion-based video compression while maintaining high perceptual quality, as shown in Fig.~\ref{fig:overview}. It stands as the first diffusion-based perceptual NVC method capable of real-time processing---achieving \textbf{206 / 30 fps} encoding/decoding speeds for 720p videos on an NVIDIA H800 GPU. Compared with prior diffusion-based NVC methods, it delivers nearly a \textbf{10$\times$} speedup, representing an important step toward practical deployment.

\section{Background}
\subsection{Neural Video Compression}
Driven by the exponential surge in video data, Neural Video Compression (NVC) has witnessed rapid progress, demonstrating substantial potential in compression efficiency. The pioneering DVC~\cite{DVC} established the first end-to-end residual coding framework by mirroring the architecture of traditional hybrid codecs. Subsequently, the DCVC family~\cite{DCVC,DCVC-TCM,DCVC-HEM,DCVC-DC,DCVC-FM,DCVC-RT} introduced a conditional coding paradigm. By extracting rich temporal contexts from previously decoded frames to condition the current frame, these methods have achieved remarkable performance across numerous benchmarks~\cite{MCL-JCV,UVG,USTC-TD}. More recent approaches~\cite{DCVC-MIP,DCVC-LCG,ECVC,DCMVC,SEVC,EHVC}, leveraging diverse context mining strategies and advanced motion modeling, have significantly outperformed the VVC standard in Low-Delay mode. Parallel efforts have also advanced NVC in Random Access mode~\cite{IBVC,B-CANF,DCVC-B,BRHVC,BiECVC}; notably, BiECVC~\cite{BiECVC} became the first NVC method to surpass VTM-RA by fully exploiting the local and non-local dependencies of bidirectional reference frames. However, these methods are predominantly optimized for pixel-level distortion metrics (e.g., MSE). Constrained by the inherent Rate-Distortion-Perception trade-off~\cite{RDP}, they frequently suffer from reconstruction blurriness and the loss of fine-grained details.

\begin{figure*}[!t]
  \vskip 0.2in
  \centering
  \centerline{\includegraphics[width=2\columnwidth]{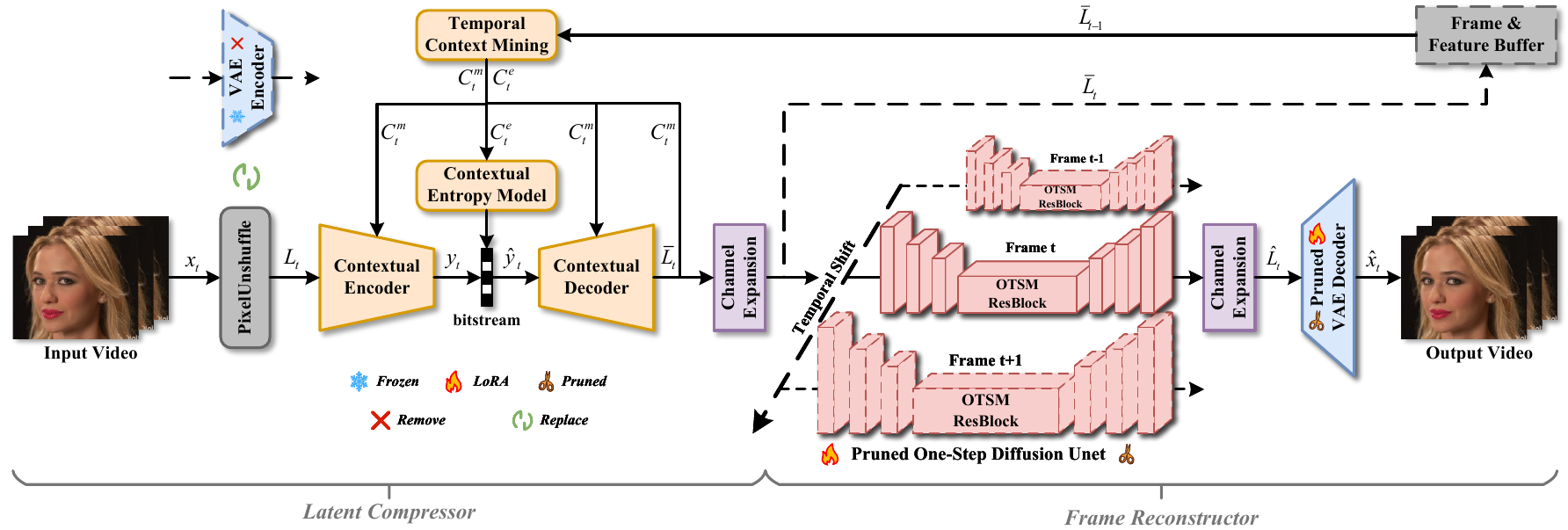}}
  \caption{
    The Efficient and Informative Model Architecture of DiffVC-RT.
  }
  \label{fig:framework}
\end{figure*}

\subsection{Diffusion-based Perceptual NVC}
To achieve visually pleasing results, Perceptual NVC has emerged as a pivotal research direction. Early works~\cite{DVC-P,PLVC,GLC-Video,GLVC} employed perceptual loss functions and Generative Adversarial Networks (GANs)~\cite{GAN} to enhance texture reconstruction; however, their efficacy was often compromised by training instability and mode collapse artifacts. Recently, Diffusion-based NVC has garnered significant attention, leveraging the robust generative priors of pre-trained foundation models to reconstruct high-fidelity video. Pioneering works such as DiffVC~\cite{DiffVC} and I2VC~\cite{I2VC} integrated these models into the video compression pipeline, synthesizing detail-rich frames via iterative multi-step denoising. Subsequently, DiffVC-OSD~\cite{DiffVC-OSD} firstly apply a one-step diffusion model to video compression, substantially reducing inference latency while enhancing perceptual quality. Similarly, recent methods like S2VC~\cite{S2VC} and YODA~\cite{YODA} have demonstrated the potential of one-step diffusion for ultra-low bitrate compression. Distinct from image-based backbones, GNVC-VD~\cite{GNVC-VD} adopts a video diffusion transformer to generate results with superior temporal consistency.

However, without exception, all diffusion-based NVC methods face the challenge of low inference efficiency due to the heavy computational burden of foundation models. This bottleneck is particularly severe in multi-step and video transformer-based approaches. While one-step diffusion has mitigated latency to some degree, it still falls significantly short of real-time industrial standards. In this work, we propose \textbf{DiffVC-RT}, a framework marking a decisive step \textit{Towards Practical Real-Time Diffusion-based Perceptual Neural Video Compression}. DiffVC-RT effectively breaks the inference efficiency bottleneck of diffusion-based NVC while maintaining competitive perceptual performance.

\section{Efficient \& Informative Model Architecture}
In this work, we present DiffVC-RT, the first real-time diffusion-based perceptual NVC framework. To navigate the trade-off among compression efficiency, perceptual quality, and inference latency and seek the optimal Pareto Frontier, we design an Efficient and Informative Model Architecture (EIMA), as illustrated in Fig.~\ref{fig:framework}.

\paragraph{Framework Overview.} DiffVC-RT is composed of two primary components:  Latent Compressor and Frame Reconstructor. 
In the \textbf{Latent Compressor}, the input video frame $x_t$ is mapped into a high-dimensional latent representation $L_t$ via the PixelUnshuffle operation. Subsequently, conditioned on temporal contexts $C^m_{t}$ and $C^e_{t}$ extracted from the previously decoded latent representation $\bar{L}_{t-1}$, $L_t$ is encoded into a bitstream and then decoded to obtain the reconstructed latent representation $\bar{L}_t$. 
In the \textbf{Frame Reconstructor}, the concatenation of $\bar{L}_t$ and $C^m_{t}$ undergoes one-step diffusion perceptual enhancement to produce $\hat{L}_t$, which is finally reconstructed into a high-perceptual-quality video frame $\hat{x}_t$ by the VAE Decoder. 
Distinct from prior work like DiffVC-OSD~\cite{DiffVC-OSD}, we discard the complex motion coding branch, adopting an architecture similar to DCVC-RT~\cite{DCVC-RT} as an efficient Latent Compressor and utilizing the lightweight pruned diffusion model proposed in AdcSR~\cite{AdcSR} as the Frame Reconstructor to minimize computational overhead. Specific details and analyses are provided below.

\paragraph{VAE Encoder Removal.} In diffusion-based NVC methods, the encoder typically employs a pre-trained VAE Encoder to compress input frames into low-dimensional, low-resolution latent representations to accommodate the subsequent diffusion process. The commonly used SD V2.1-base~\cite{SD} compresses a $3 \times H \times W$ frame into a $4 \times \frac{H}{8} \times \frac{W}{8}$ latent, resulting in severe loss of structural information. Furthermore, the VAE Encoder, with its progressive downsampling, dominates the computational cost at the encoder side. Inspired by DCVC-RT~\cite{DCVC-RT}, we remove the VAE Encoder and directly adopt the \textbf{PixelUnshuffle} operation for space-to-depth conversion. By contrast, PixelUnshuffle preserves all high-frequency details of the input frame with zero information loss and incurs negligible computational cost. This improvement completely offloads the computational burden of the encoder to the efficient Latent Compressor, establishing DiffVC-RT as the first diffusion-based NVC framework capable of real-time encoding for 1080p video.

\paragraph{Latent Channel Expansion.} The input and output channel dimensions of the U-Net and VAE Decoder in most foundation diffusion models are fixed at 4. This creates a significant dimensional mismatch between the high-dimensional features output by the Latent Compressor and the Frame Reconstructor. Forcing dimensionality reduction leads to substantial attenuation of semantic and texture information in the compressed features. To overcome this bottleneck, we restructure the input and output interfaces of the U-Net and VAE Decoder, expanding the latent dimension from 4 to 256 to align with the Latent Compressor. This construction establishes an end-to-end \textbf{``High-throughput Highway''} at a minimal computational cost, ensuring the flow of information without any passive loss.

\paragraph{Pruned U-Net \& VAE Decoder.} For the decoder side, the computational bottleneck lies primarily in the massive U-Net and the progressively upsampling VAE Decoder. To accelerate decoding speed, we initialize our Frame Reconstructor using the pre-trained Pruned U-Net and VAE Decoder from AdcSR~\cite{AdcSR}. Specifically, the \textbf{Pruned U-Net} discards text-prompt and time-embedding modules that have negligible impact on reconstruction, while pruning intermediate feature channels by 25\%. Similarly, the \textbf{Pruned VAE Decoder} reduces intermediate channels by 50\%. Furthermore, regarding the injection of temporal context, we eschew the heavy ControlNet~\cite{ControlNet} employed in DiffVC~\cite{DiffVC} or the computationally intensive Temporal Context Adapter used in DiffVC-OSD~\cite{DiffVC-OSD}. Instead, we adopt a streamlined strategy by simply concatenating $\bar{L}_t$ and $C^m_t$ to explicitly incorporate information from previous frames into the U-Net. By leveraging these lightweight architectures and this efficient temporal conditioning, we significantly accelerate decoding speed and reduce computational overhead while maintaining powerful generative priors.

\section{Explicit \& Implicit Consistency Modeling}
Beyond severe information loss and prohibitive inference latency, diffusion-based NVCs face a critical challenge: temporal inconsistency arising from the inherent stochasticity of generative models. Previous works employed computationally expensive consistency modeling modules (e.g., TCG in S2VC \cite{S2VC}, TA-AE in YODA \cite{YODA}, or even 3D VAE and video diffusion transformer in GNVC-VD \cite{GNVC-VD}) to enhance the temporal stability. However, these approaches incur a substantial computational burden. To strictly align with our goal of real-time inference, we propose a \textbf{``Zero-Cost''} Explicit and Implicit Consistency Modeling method.

\subsection{Explicit Consistency Enhancement}
In diffusion-based NVC, the root cause of temporal inconsistency lies in the independent diffusion of each frame, lacking sufficient inter-frame interaction. Although DiffVC \cite{DiffVC} and DiffVC-OSD \cite{DiffVC-OSD} use temporal context as diffusion conditions, their modulation frequency is insufficient. While GNVC-VD \cite{GNVC-VD}, S2VC \cite{S2VC}, and YODA \cite{YODA} achieve high-intensity interaction, they inevitably introduce significant computational complexity. Therefore, we propose a high-frequency, high-intensity, and high-efficiency method for inter-frame interaction. Specifically, we insert the \textbf{Online Temporal Shift Module (OTSM)} \cite{TSM} into the ResBlocks of the U-Net. This module shifts a proportion ($1/P$) of the feature channels by $+1$ along the temporal dimension, enabling the diffusion enhancement process of the current frame to explicitly incorporate information from the corresponding position of the previous frame. Benefiting from the substantial number of ResBlocks in the U-Net and explicit temporal shift, our proposal achieves high-frequency and high-intensity inter-frame interaction within a single diffusion step. Notably, this temporal shift operation is \textbf{almost cost-free}, introducing neither computational complexity nor parameters, fully aligning with our real-time inference objective.

\subsection{Implicit Consistency Constraints}
To further enhance the temporal consistency of reconstructed videos without introducing additional inference overhead, we propose hybrid Implicit Consistency Constraints. Operating from the loss function perspective, these constraints utilize motion priors between adjacent frames to penalize non-physical temporal flickering and jittering. This formulation consists of two components: Pixel Warping Loss and Feature Warping Loss.

\textbf{Pixel Warping Loss.} This loss is based on the assumption of short-term temporal coherence: the content of the current frame $x_t$ should be deducible from the previous frame $x_{t-1}$ via optical flow. Specifically, we use a pre-trained optical flow network, RAFT-lite \cite{RAFT}, to estimate the flow field $v_{t \to t-1}$ from $x_t$ to $x_{t-1}$, and compute an occlusion mask $m_t$ to exclude interference from occluded regions. The pixel warping loss is defined as:
\begin{equation}
  \mathcal{L}_{p} = \frac{\sum m_t \odot \lVert x_t - \mathcal{W}(x_{t-1}, v_{t \to t-1}) \rVert_1}{\sum m_t}
\end{equation}
where $\mathcal{W}(\cdot)$ denotes the backward warping and $\odot$ represents element-wise multiplication. While effective for accurate region alignment, $\mathcal{L}_p$ is overly rigid and prone to blurring or ghosting under flow errors or object deformations.

\textbf{Feature Warping Loss.} To mitigate the blurring issues caused by pixel-level constraints and ensure the temporal stability of structural semantics, we further introduce Feature Warping Loss. Unlike the pixel domain, deep feature space offers larger receptive fields and greater semantic robustness, tolerating minor misalignments while enforcing natural texture translation along motion trajectories. We extract feature maps $\phi^l(\cdot)$ at the \textit{relu3\_4} layer by feeding $x_t$ and $x_{t-1}$ into a pre-trained VGG \cite{VGG}. To match the spatial resolution, we downsample and scale the optical flow $v_{t \to t-1}$ and mask $m_t$ to obtain $v^l_{t \to t-1}$ and $m^l_t$. The feature warping loss is defined as:
\begin{equation}
  \mathcal{L}_{f} = \frac{\sum m^{l}_t \odot \lVert \phi^l(x_t) - \mathcal{W}(\phi^l(x_{t-1}), v^{l}_{t \to t-1}) \rVert_2^2}{\sum m^{l}_t}
\end{equation}
This loss enforces coherence on the deep feature manifold, effectively suppressing texture flickering and complementing the pixel-level constraint.

The final hybrid implicit consistency constraints is $\mathcal{L}_{ICC} = \lambda_{p} \mathcal{L}_{p} + \lambda_{f} \mathcal{L}_{f}$. $\mathcal{L}_{ICC}$ is a training-only objective, and all auxiliary modules are discarded at inference, resulting in \textbf{fully cost-free} Implicit Consistency Constraints.

\section{Asynchronous \& Parallel Decoding Pipeline}
\begin{figure}[!t]
  \vskip 0.2in
  \centering
  \centerline{\includegraphics[width=\columnwidth]{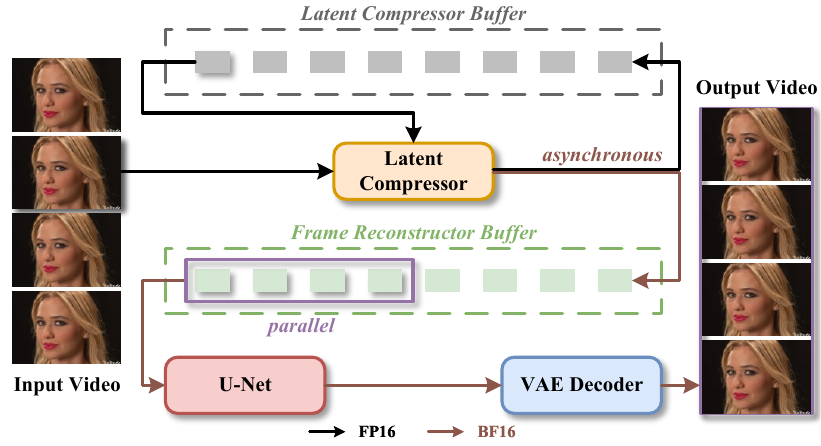}}
  \caption{
    The Asynchronous and Parallel Decoding Pipeline.
  }
  \label{fig:APDP}
\end{figure}

To bridge the final gap toward real-time decoding, we propose a novel efficient decoding scheme tailored for DiffVC-RT: the Asynchronous and Parallel Decoding Pipeline, as shown in Fig.~\ref{fig:APDP}. With this scheme, we significantly accelerate decoding speed with negligible performance impact, establishing DiffVC-RT as the first diffusion-based NVC method to support real-time decoding of 720p videos.

\subsection{Asynchronous and Parallel Decoding}
\paragraph{Asynchronous Decoding.} As illustrated in Fig.~\ref{fig:APDP}, in DiffVC-RT, the coding of the next frame depends solely on the reconstructed latent representation of the previous frame, $\bar{L}_{t-1}$. This implies that the Latent Compressor operates within the prediction loop (In-Loop), whereas the Frame Reconstructor operates outside the loop (Out-of-Loop). Leveraging this property, we propose to \textbf{decouple the execution of Latent Decoding and Frame Reconstruction asynchronously.} Specifically, we maintain independent buffers for the Latent Compressor and the Frame Reconstructor. The former stores $\bar{L}_t$ required for the next frame's latent decoding, while the latter stores $\bar{L}_t$ and $C^m_t$ required for the current frame's reconstruction. These two processes execute concurrently without interference, thereby enhancing overall decoding efficiency.

\paragraph{Parallel Frame Reconstruction.} Since the Frame Reconstructor is computationally more complex than the Latent Compressor, a simple asynchronous strategy would result in a speed imbalance between the two branches, limiting the overall efficiency gain. To address this bottleneck, we propose Parallel Frame Reconstruction. Specifically, we fetch $N$ samples from the Frame Reconstructor Buffer at a time, concatenate them along the batch dimension, and pass them through the U-Net and VAE Decoder in parallel. However, this strategy conflicts with the previously OTSM in Explicit Consistency Enhancement, which relies on sequential frame dependency. To resolve this, we introduce the \textbf{Batch-dimension OTSM}. This module performs a $+1$ shift on partial channels along the batch dimension within a batch (Intra-batch Shift), and passes the partial channels of the last sample in the current batch to the subsequent batch (Inter-batch Shift). In summary, the Parallel Frame Reconstruction strategy achieves a breakthrough in decoding speed at the cost of an $N-1$ frame latency, which is acceptable in most real-time scenarios~\cite{UI2C}.

\begin{figure*}[!t]
  \vskip 0.2in
  \centering
  \centerline{\includegraphics[width=2\columnwidth]{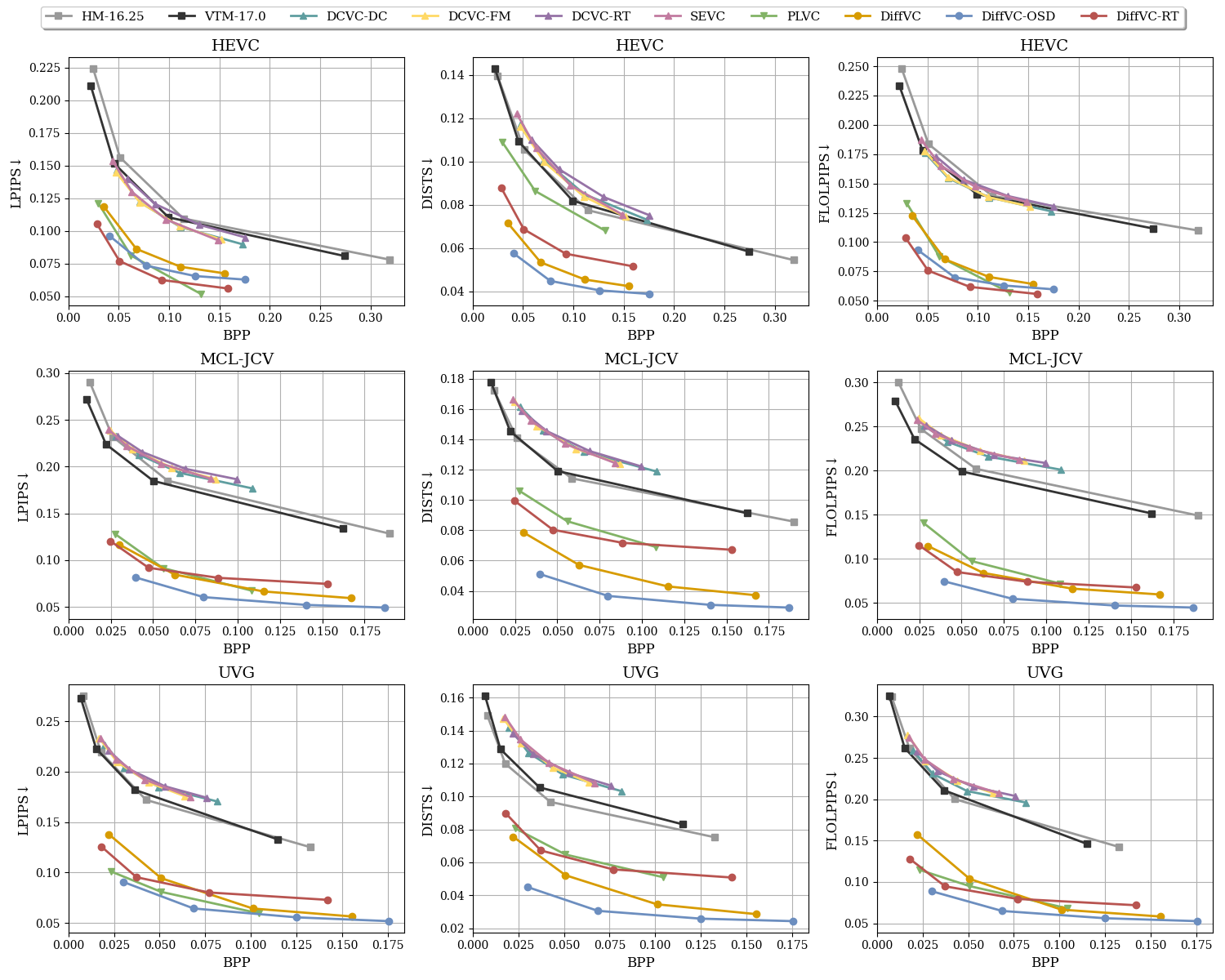}}
  \caption{
    Rate-perception curves of the proposed DiffVC-RT and other methods on the HEVC, MCL-JCV, and UVG datasets.
  }
  \label{fig:main_rd_curve}
\end{figure*}

\subsection{Mixed Half Precision}
Previous works~\cite{DCVC-FM,DCVC-RT} have adopted half-precision inference strategies, delivering speedups with negligible performance degradation. However, when applying standard half-precision format (FP16) to DiffVC-RT, numerical overflow occurs during the forward pass of the U-Net and VAE Decoder due to the limited dynamic range of FP16. To resolve this issue while maintaining low latency, we propose a Mixed Half Precision strategy. We employ the \textbf{BFloat16 (BF16)} format for the inference of the dynamic-range-sensitive U-Net and VAE Decoder, while the precision-sensitive Latent Compressor remains in FP16. BF16 shares the same exponent width as FP32, preventing precision overflow, while reducing memory bandwidth by 50\% and significantly increasing computational throughput.

\section{Experiments}
\subsection{Experimental Setup}
\paragraph{Datasets.} For training, we utilize the Vimeo-90k~\cite{Vimeo90k} dataset. For evaluation, we employ the HEVC (Class B$\sim$E)~\cite{HEVC}, UVG~\cite{UVG}, and MCL-JCV~\cite{MCL-JCV} datasets, which encompass diverse resolutions and scene complexities.

\paragraph{Compared Methods.} We conduct a comprehensive comparison of the proposed \textbf{DiffVC-RT} with various established video compression methods. These include traditional reference software: HM-16.25~\cite{HEVC} and VTM-17.0~\cite{VVC}; Distortion-oriented NVC methods: DCVC-DC~\cite{DCVC-DC}, DCVC-FM~\cite{DCVC-FM}, DCVC-RT~\cite{DCVC-RT}, and SEVC~\cite{SEVC}; GAN-based perceptual NVC methods: PLVC~\cite{PLVC}; and diffusion-based perceptual NVC methods: DiffVC~\cite{DiffVC} and DiffVC-OSD~\cite{DiffVC-OSD}. For recent diffusion-based works without open-source implementations, such as GNVC-VD~\cite{GNVC-VD}, S2VC~\cite{S2VC}, and YODA~\cite{YODA}, we extract results directly from their original papers to ensure the fairest possible comparison.

\paragraph{Training Settings.} We employ a seven-stage training strategy. The first six stages train specific modules individually using customized loss functions, followed by a final stage of joint fine-tuning to achieve optimal performance. The Adam optimizer~\cite{Adam} is used with $\beta_1 = 0.9$ and $\beta_2 = 0.999$. Similar to the variable bitrate scheme in DCVC-FM~\cite{DCVC-FM}, we define 16 bitrate points for DiffVC-RT. In each training step, a Lagrange multiplier $\lambda$ is randomly selected from the range $[16, 384]$ to balance bitrate and reconstruction quality. Further training details and parameter settings are provided in the Appendix~\ref{appendix:train}.

\begin{table}[t]
  \caption{Speed analysis. The encoding / decoding speed (measured in frames per second, fps) are evaluated across various resolutions, methods and devices. ``OOM'' indicates out-of-memory conditions. ``-'' denotes data is unavailable. $^{*}$ denotes device conditions and data reported in the original publication.}
  \label{tab:speed}
  \centering
  \scriptsize    
  \setlength{\tabcolsep}{1.5pt}
  \renewcommand{\arraystretch}{1.05}

  \begin{tabular*}{\columnwidth}{@{\extracolsep{\fill}}lcccc@{}}
  \toprule
  \textbf{Method} & \textbf{416$\times$240} & \textbf{832$\times$480} & \textbf{1280$\times$720} & \textbf{1920$\times$1080} \\ \midrule
  \multicolumn{5}{@{}l@{}}{\textcolor{gray}{\emph{AMD EPYC 7J13 processor}}} \\
  HM-16.25      & 0.488 / 216.37 & 0.116 / 56.70 & 0.092 / 37.17 & 0.027 / 12.03 \\
  VTM-17.0      & 0.060 / 122.88 & 0.014 / 34.80 & 0.025 / 27.22 & 0.004 / 8.04  \\ \midrule
  \multicolumn{5}{@{}l@{}}{\textcolor{gray}{\emph{NVIDIA GeForce RTX 3090 (24GB PCIe 350W)}}} \\ 
  DCVC-DC       & 22.54 / 25.48 & 8.86 / 9.96 & 4.01 / 5.01 & 1.76 / 2.25 \\
  DCVC-FM       & 21.71 / 25.05 & 9.07 / 9.73 & 4.17 / 4.84 & 1.83 / 2.20 \\
  DCVC-FM(half) & 21.57 / 25.45 & 12.71 / 14.01 & 6.95 / 7.26 & 2.99 / 3.39 \\
  DCVC-RT       & 66.97 / 64.21 & 44.61 / 53.68 & 29.02 / 29.50 & 14.62 / 13.86 \\
  DCVC-RT(half) & 153.02 / 166.52 & 128.62 / 133.93 & 97.52 / 95.55 & 46.54 / 48.82 \\
  DiffVC        & 0.60 / 0.63 & 0.22 / 0.23 & 0.07 / 0.07 & OOM / OOM \\
  DiffVC-OSD    & 5.12 / 5.96 & 2.16 / 2.78 & 0.85 / 1.11 & 0.31 / 0.40 \\
  \textbf{DiffVC-RT} & \textbf{159.09 / 55.89} & \textbf{134.74 / 21.11} & \textbf{92.62 / 8.75} & \textbf{46.03 / 3.13} \\ \midrule
  \multicolumn{5}{@{}l@{}}{\textcolor{gray}{\emph{NVIDIA A800 (80GB PCIe 300W)}}} \\ 
  DCVC-DC       & 36.24 / 43.88 & 15.63 / 18.71 & 7.47 / 9.36 & 3.43 / 4.36 \\
  DCVC-FM       & 36.45 / 42.40 & 16.20 / 17.97 & 7.59 / 8.96 & 3.48 / 4.20 \\
  DCVC-FM(half) & 36.86 / 43.52 & 20.89 / 23.54 & 10.65 / 12.31 & 4.96 / 5.94 \\
  DCVC-RT       & 112.61 / 127.81 & 97.20 / 102.85 & 68.81 / 65.01 & 32.06 / 29.92 \\
  DCVC-RT(half) & 237.65 / 249.62 & 218.33 / 233.66 & 205.28 / 190.06 & 111.53 / 105.40 \\
  DiffVC        & 0.85 / 0.88 & 0.29 / 0.29 & 0.09 / 0.09 & 0.02 / 0.02 \\
  DiffVC-OSD    & 7.60 / 8.85 & 4.26 / 5.47 & 1.76 / 2.33 & 0.67 / 0.87 \\
  \textbf{DiffVC-RT} & \textbf{240.84 / 130.83} & \textbf{225.53 / 54.47} & \textbf{189.65 / 21.19} & \textbf{105.94 / 7.67} \\ \midrule
  \multicolumn{5}{@{}l@{}}{\textcolor{gray}{\emph{NVIDIA H800 (80GB PCIe 350W)}}} \\ 
  DCVC-DC       & 41.22 / 47.09 & 18.53 / 21.15 & 8.49 / 10.82 & 3.88 / 5.04 \\
  DCVC-FM       & 41.72 / 47.25 & 19.11 / 20.83 & 8.78 / 10.49 & 4.00 / 4.88 \\
  DCVC-FM(half) & 43.21 / 50.40 & 25.50 / 27.07 & 12.78 / 14.55 & 5.86 / 6.98 \\
  DCVC-RT       & 133.57 / 150.00 & 113.08 / 123.91 & 80.22 / 78.01 & 35.32 / 35.81 \\
  DCVC-RT(half) & 273.87 / 287.34 & 231.19 / 245.73 & 213.06 / 214.56 & 139.37 / 130.92 \\
  DiffVC        & 0.96 / 0.99 & 0.47 / 0.48 & 0.15 / 0.15 & 0.04 / 0.04 \\
  DiffVC-OSD    & 9.95 / 11.26 & 5.42 / 6.88 & 2.33 / 3.10 & 0.90 / 1.18 \\
  \textbf{DiffVC-RT}     & \textbf{275.97 / 154.79} & \textbf{244.20 / 74.79} & \textbf{206.00 / 30.02} & \textbf{138.11 / 10.95} \\ \midrule
  \multicolumn{5}{@{}l@{}}{\textcolor{gray}{\emph{NVIDIA A800$^{*}$}}} \\ 
  GNVC-VD       & - / - & - / - & - / - & 6.54 / 0.64 \\ \midrule
  \multicolumn{5}{@{}l@{}}{\textcolor{gray}{\emph{NVIDIA A100$^{*}$}}} \\ 
  S2VC          & - / - & - / - & - / - & 6.60 / 1.27 \\ \midrule
  \multicolumn{5}{@{}l@{}}{\textcolor{gray}{\emph{NVIDIA GeForce RTX 5090$^{*}$}}} \\ 
  YODA          & - / - & - / - & - / - & - / 0.97 \\ \bottomrule
  \end{tabular*}
\end{table}

\paragraph{Evaluation Settings.} Following previous works~\cite{DCVC-DC,DiffVC,DiffVC-OSD}, we evaluate the first 96 frames of each video sequence, with the intra period set to 32. The low-delay encoding configuration is adopted, and the efficient PO-RTIntra~\cite{PO-RTIntra} is employed to encode I-frames. Perceptual quality is evaluated using LPIPS~\cite{LPIPS} and DISTS~\cite{DISTS}, with FloLPIPS~\cite{FloLPIPS} additionally reported to measure temporal perceptual consistency. Distortion quality is assessed using PSNR and MS-SSIM~\cite{MS-SSIM}.

\paragraph{Implementation Details.} Regarding the Diffusion Model, we initialize the U-Net and VAE Decoder using pre-trained weights from AdcSR~\cite{AdcSR} and fine-tune them using LoRA~\cite{LoRA}, with ranks set to 48 and 16, respectively. For the OTSM in Explicit Consistency Enhancement, the temporal shift channel ratio $1/P$ is set to the default configuration of $P=8$. For Implicit Consistency Constraints, we set $\lambda_p=1.0$ and $\lambda_f=2 \times 10^{-4}$. Regarding the parallel number $N$ in the Parallel Frame Reconstruction strategy, we set $N=8$ for GPUs with large memory capacity (e.g., NVIDIA H800 and A800) and $N=4$ for GPUs with limited memory capacity (e.g., RTX 3090).

\begin{table}[t]
  \caption{Complexity analysis. The computational complexity and parameter count are reported. ``-'' denotes data is unavailable.}
  \label{tab:complexity}
  \centering
  \scriptsize   
  \setlength{\tabcolsep}{4pt}
  \renewcommand{\arraystretch}{1.05}

  \begin{tabular*}{\columnwidth}{@{\extracolsep{\fill}}lcc@{}}
  \toprule
  \textbf{Method}     & \textbf{kMACs/pixel} & \textbf{Params (M)} \\ \midrule
  DCVC-DC             & 1335.58 & 19.80 \\
  DCVC-FM             & 1128.62 & 18.34 \\
  DCVC-RT             & 194.94 & 20.69 \\
  DiffVC              & 110083.01 & 1686.07 \\
  DiffVC-OSD          & 9393.59 & 1488.67 \\
  GNVC-VD             & - & 2334.50 \\
  S2VC                & - & 1346.00 \\
  YODA                & - & 1063.15 \\
  \textbf{DiffVC-RT}  & \textbf{2462.36} & \textbf{474.26} \\ \bottomrule
  \end{tabular*}
\end{table}

\begin{figure}[!t]
  \vskip 0.2in
  \centering
  \centerline{\includegraphics[width=\columnwidth]{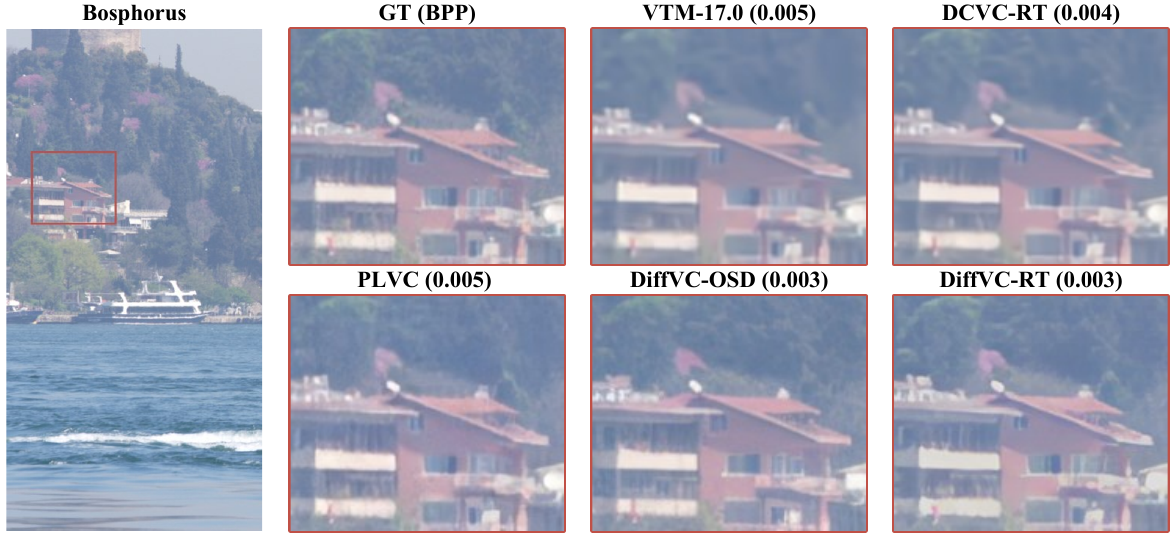}}
  \caption{
    Visual comparison across different codecs.
  }
  \label{fig:main_visual_result}
\end{figure}

\subsection{Main Results}
\paragraph{Performance Comparison.} Fig.~\ref{fig:main_rd_curve} illustrates the rate-perception curves of DiffVC-RT and other video codecs across three benchmark datasets. DiffVC-RT significantly outperforms traditional codecs and distortion-oriented NVCs across all perceptual metrics and datasets. Among perceptual NVC methods, DiffVC-RT also demonstrates highly competitive performance. Particularly on the HEVC dataset, DiffVC-RT achieves State-of-the-Art performance in both LPIPS and FloLPIPS metrics, surpassing the previous best method, DiffVC-OSD~\cite{DiffVC-OSD}. For more specific BD-Rate and BD-Metric values, as well as performance on distortion metrics, please refer to Appendix~\ref{appendix:result}.

\paragraph{Visual Comparison.} Fig.~\ref{fig:main_visual_result} presents a qualitative comparison of reconstructed frames. DiffVC-RT demonstrates the capability to synthesize detail-rich, high-fidelity textures while maintaining efficient inference efficiency. Extended visualization results are provided in Appendix~\ref{appendix:visual}.

\begin{table*}[!t]
  \caption{Ablation studies of DiffVC-RT. All performance results are tested BD-Rate ($\%$) $\downarrow$ on HEVC Class C with DiffVC-RT as the anchor. All fps results are tested on 832$\times$480 videos with an NVIDIA H800 GPU.}
  \label{tab:ablation}
  \centering
  \scriptsize  
  \resizebox{1.0\textwidth}{!}{
  \begin{tabular}{@{}lcccccccc@{}}
  \toprule
  \multirow{2}{*}{\textbf{Evaluative Dimension}} & \multicolumn{5}{c}{\textbf{Performance}} & \multicolumn{3}{c}{\textbf{Efficiency}} \\
  \cmidrule(lr){2-5}\cmidrule(lr){6-9}
  & \textbf{LPIPS} & \textbf{DISTS} & \textbf{FloLPIPS} & \textbf{Average} & \textbf{Params (M)} & \textbf{kMACs/pixel} & \textbf{Enc. fps} & \textbf{Dec. fps} \\ \midrule
  \multicolumn{8}{@{}l@{}}{\textcolor{gray}{\emph{Baseline}}} \\ 
  \textbf{DiffVC-OSD} & 57.51 & -51.06 & 51.96 & 19.47 & 1488.67 & 9393.59 & 5.42 & 6.88 \\ \midrule
  \multicolumn{8}{@{}l@{}}{\textcolor{gray}{\emph{Efficient \& Informative Model Architecture (EIMA)}}} \\ 
  \textbf{A}: base & 49.97 & -45.54 & 33.23 & 12.55 & 1386.45 & 9152.45 & 22.66 & 7.64 \\
  \textbf{B}: A + Pruned U-Net \&  VAE Decoder & 27.48 & -47.96 & 28.24 & 2.59 & 508.24 & 4652.78 & 22.60 & 13.96 \\
  \textbf{C}: B + VAE Encoder Removement & 32.16 & 17.18 & 31.77 & 27.04 & 473.99 & 2458.14 & 101.60 & 14.24 \\
  \textbf{D}: C + Latent Channel Expansion & 11.60 & 53.60 & 14.84 & 26.68 & 474.26 & 2462.36 & 108.14 & 14.50 \\ \midrule
  \multicolumn{8}{@{}l@{}}{\textcolor{gray}{\emph{Explicit \& Implicit Consistency Modeling (EICM)}}} \\ 
  \textbf{E}: D + Explicit Consistency Enhancement & 5.48 & 26.71 & 7.56 & 13.25 & 474.26 & 2462.36 & 105.75 & 14.36 \\
  \textbf{F}: E + Implicit Consistency Constraints & 3.34 & 2.91 & 7.47 & 4.57 & 474.26 & 2462.36 & 106.88 & 14.35 \\
  \textbf{G}: F + End-to-End Finetune & 0.77 & 0.82 & 0.37 & 0.65 & 474.26 & 2462.36 & 108.45 & 14.47 \\ \midrule
  \multicolumn{8}{@{}l@{}}{\textcolor{gray}{\emph{Asynchronous \& Parallel Decoding Pipeline (APDP)}}} \\ 
  \textbf{H}: G + Mixed Half Precision & 0.00 & 0.00 & 0.00 & 0.00 & 474.26 & 2462.36 & 242.97 & 23.75 \\
  \textbf{DiffVC-RT}: H + Asynchronous Parallel Pipeline & 0.00 & 0.00 & 0.00 & 0.00 & 474.26 & 2462.36 & 244.20 & 74.79 \\ \bottomrule
  \end{tabular}
  }
\end{table*}

\paragraph{Speed Analysis.} Table~\ref{tab:speed} reports the encoding and decoding speed across varying resolutions and devices. To ensure a fair comparison, we also include half-precision inference results for DCVC-FM~\cite{DCVC-FM} and DCVC-RT~\cite{DCVC-RT}. On the \textbf{encoder} side, benefiting from our streamlined architecture where complexity is confined to the efficient Latent Compressor, DiffVC-RT achieves real-time speed across all tested devices and resolutions, matching the efficiency of DCVC-RT~\cite{DCVC-RT}. On the \textbf{decoder} side, DiffVC-RT achieves a remarkable speedup of approximately $10\times$ over DiffVC-OSD~\cite{DiffVC-OSD} and two orders of magnitude ($100\times$) over DiffVC~\cite{DiffVC}. Crucially, it stands as the first diffusion-based NVC method capable of real-time decoding ($>30$ fps) 720p videos on an NVIDIA H800 GPU. Furthermore, compared to recent concurrent works such as GNVC-VD~\cite{GNVC-VD}, S2VC~\cite{S2VC}, and YODA~\cite{YODA}, DiffVC-RT shows a significant advantage in inference latency based on reported figures.

\paragraph{Complexity Analysis.} Table~\ref{tab:complexity} summarizes the computational complexity and parameter count of various video codecs, profiled using the DeepSpeed library. Benefiting from the utilization of a lightweight pruned Diffusion Model, DiffVC-RT achieves the lowest parameter count and computational overhead among all diffusion-based NVC methods.

\subsection{Ablation Studies}
To validate the efficacy of the proposed DiffVC-RT design, we conducted comprehensive ablation studies. As summarized in Table~\ref{tab:ablation}, we report the BD-Rate performance (on perceptual metrics) and efficiency profiles (parameters, complexity, and speed) for various model configurations.

Starting from DiffVC-OSD~\cite{DiffVC-OSD} as the baseline, we obtain Model A by removing the motion branch in the latent compressor, improving both perceptual quality and efficiency.
For \textbf{EIMA}, adopting the pruned U-Net and VAE decoder yields consistent gains. Replacing the VAE encoder with PixelUnshuffle plus Latent Channel Expansion is key to real-time encoding and structural information preservation, but degrades perceptual performance due to the absence of generative priors from the pretrained VAE encoder, considering a quality-efficiency trade-off.
In terms of \textbf{EICM}, Explicit Consistency Enhancement yields a 13.43\% improvement via zero-overhead inter-frame interaction. Implicit Consistency Constraints add an 8.68\% gain via temporal penalty in pixel and feature spaces. Moreover, End-to-End Joint Fine-tuning amplifies the explicit and implicit consistency modeling capabilities, yielding an additional 3.92\% boost.
Finally, for \textbf{APDP}, the Mixed Half Precision strategy significantly accelerates inference with negligible performance impact. The Asynchronous Parallel Pipeline further unlocks a $3\times$ speedup in the decoding process. Please refer to Appendix~\ref{appendix:ablation} for a detailed analysis.

\section{Conclusion}
In this paper, we have proposed \textbf{DiffVC-RT}, the first framework designed for real-time diffusion-based perceptual NVC. Through an Efficient and Informative Model Architecture, DiffVC-RT establishes a lightweight and computationally efficient paradigm while mitigating structural information loss. Furthermore, by employing a ``Zero-Cost'' Explicit and Implicit Consistency Modeling strategy, we effectively suppress temporal flickering in reconstructed videos without incurring additional overhead. Finally, leveraged by our tailored Asynchronous and Parallel Decoding Pipeline, DiffVC-RT achieves a breakthrough in inference latency. To the best of our knowledge, DiffVC-RT is the first diffusion-based perceptual NVC method capable of real-time encoding for 1080p videos and real-time decoding for 720p videos, while maintaining competitive perceptual performance. This represents a significant breakthrough for diffusion-based perceptual NVC frameworks.


\bibliography{reference}
\bibliographystyle{icml2026}

\newpage
\appendix
\onecolumn
\section{Model Architecture}\label{appendix:model}
\begin{figure*}[!h]
  \vskip 0.2in
  \centering
  \centerline{\includegraphics[width=0.5\columnwidth]{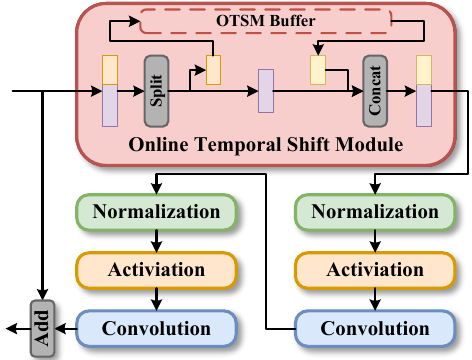}}
  \caption{
    Detailed architecture of the ResBlock with the Online Temporal Shift Module in the U-Net.
  }
  \label{fig:otsm_resblock}
\end{figure*}

This section provides supplementary details regarding the model architecture of DiffVC-RT. The architecture of the \textbf{Latent Compressor} is similar to that of DCVC-RT~\cite{DCVC-RT}, with the specific exception that the reconstruction module is removed. For the generative backbone, Fig.~\ref{fig:otsm_resblock} illustrates the structure of the \textbf{OTSM ResBlock} embedded within the U-Net. Guided by prior findings~\cite{TSM} that favor residual insertion over main-path placement, we integrate the OTSM into the residual branches of all 22 ResBlocks. Specifically, input features are partitioned along the channel dimension at a ratio of $1 : (P-1)$; the first segment is cached in the OTSM Buffer for the subsequent frame, while the remaining segment is concatenated with the buffered feature slice from the previous frame. This mechanism effectively achieves explicit inter-frame interaction with zero computational overhead.

\section{Training Strategy}\label{appendix:train}
To ensure the robust optimization of DiffVC-RT, we employ a comprehensive seven-stage training curriculum. The specific configurations and hyper-parameters for each phase are detailed in Table~\ref{tab:training_strategy}.

Regarding the abbreviation, LC, VBP, and FR denote the Latent Compressor, Variable Bitrate Parameters, and Frame Reconstructor, respectively. The loss objectives are abbreviated as follows: $\mathcal{L}_D$ (Distortion), $\mathcal{L}_{RD}$ (Rate-Distortion), $\mathcal{L}_{RDP}$ (Rate-Distortion-Perceptual), $\mathcal{L}_{DPT}$ (Distortion-Perceptual-Temporal), and $\mathcal{L}_{RDPT}$ (Rate-Distortion-Perceptual-Temporal). Furthermore, $w_t$ corresponds to the hierarchical temporal weights introduced in prior work~\cite{DCVC-DC}, initialized as $[0.5, 1.2, 0.5, 0.9]$. ICC refers to our proposed Implicit Consistency Constraints, and $\mathbb{E}_t$ signifies averaging across the temporal dimension.

In summary, this meticulously designed progressive training strategy enables DiffVC-RT to achieve competitive compression performance while maintaining superior computational efficiency.

\begin{table}[h]
  \centering
  \scriptsize 
  \renewcommand{\arraystretch}{1.3} 
  \setlength{\tabcolsep}{3pt} 
  \caption{Detailed training strategy and hyper-parameter settings across 7 stages.}
  \label{tab:training_strategy}
  
  \begin{tabular*}{\linewidth}{@{\extracolsep{\fill}} c c c c c c c c c c l c c @{}}
      \toprule
      \textbf{Stage} & \textbf{\makecell{Epoch\\Num}} & \textbf{\makecell{Batch\\Size}} & \textbf{\makecell{Patch\\Size}} & \textbf{\makecell{Frame\\Num}} & \textbf{\makecell{Learning\\Rate}} & \textbf{\makecell{Trainable\\Modules}} & \textbf{\makecell{Rate\\Point}} & \textbf{$\lambda$} & \textbf{Loss} & \textbf{Loss Formula} & \textbf{\makecell{Comperss\\I-Frame}} & \textbf{\makecell{Cascade\\Train}} \\
      \midrule
      
      \multirow{5}{*}{1} & 1 & \multirow{5}{*}{48} & \multirow{5}{*}{$256\times 256$} & 2 & \multirow{5}{*}{$10^{-4}$} & \multirow{5}{*}{LC} & \multirow{5}{*}{15} & \multirow{5}{*}{384} & \multirow{5}{*}{$\mathcal{L}_D$} & \multirow{5}{*}{$\lambda(1.0 \text{MSE})$} & \multirow{5}{*}{\xmark} & \multirow{5}{*}{\xmark} \\
      & 1 & & & 3 & & & & & & & & \\
      & 2 & & & 4 & & & & & & & & \\
      & 2 & & & 5 & & & & & & & & \\
      & 4 & & & 7 & & & & & & & & \\
      \midrule
      
      2 & 10 & 48 & $256\times 256$ & 7 & $10^{-4}$ & LC & 15 & 384 & $\mathcal{L}_{RD}$ & $R + \lambda(1.0 \text{MSE})$ & \xmark & \xmark \\
      \midrule
      
      3 & 20 & 48 & $256\times 256$ & 7 & \makecell{$10^{-4}$\\$\downarrow$\\$5\cdot10^{-6}$} & \makecell{LC + VBP} & 0-15 & \makecell{16\\$\sim$\\384} & $\mathcal{L}_{RD}$ & $R + \lambda w_t(1.0 \text{MSE})$ & \cmark & \xmark \\
      \midrule
      
      4 & 30 & 48 & $256\times 256$ & 7 & \makecell{$5\cdot10^{-5}$\\$\downarrow$\\$10^{-6}$} & \makecell{LC + VBP} & 0-15 & \makecell{16\\$\sim$\\384} & $\mathcal{L}_{RDP}$ & \makecell[l]{$R + \lambda w_t(0.5 \text{MSE}$ \\ $+ 0.025 \text{LPIPS}$ \\ $+ 0.025 \text{DISTS})$} & \cmark & \xmark \\
      \midrule
      
      5 & 2 & 12 & $256\times 256$ & 7 & $10^{-5}$ & \makecell{LC + VBP} & 0-15 & \makecell{16\\$\sim$\\384} & $\mathcal{L}_{RDP}$ & \makecell[l]{$\mathbb{E}_t [ R + \lambda w_t(0.5 \text{MSE}$ \\ $+ 0.025 \text{LPIPS}$ \\ $+ 0.025 \text{DISTS}) ]$} & \cmark & \cmark \\
      \midrule
      
      \multirow{2}{*}{6} & 16 & 48 & \multirow{2}{*}{$448 \times 256$} & 7 & \makecell{$3\cdot10^{-4}$\\$\downarrow$\\$10^{-5}$} & \multirow{2}{*}{FR} & \multirow{2}{*}{0-15} & \multirow{2}{*}{\makecell{16\\$\sim$\\384}} & \multirow{2}{*}{$\mathcal{L}_{DPT}$} & \makecell[l]{$2.0 \text{MSE} + 1.0 \text{LPIPS}$ \\ $+ 1.0 \text{DISTS} + 0.1 \text{ICC}$} & \cmark & \xmark \\
       & 6 & 14 & & 5 & $5\cdot10^{-5}$ & & & & & \makecell[l]{$\mathbb{E}_t [ 2.0 \text{MSE} + 1.0 \text{LPIPS}$ \\ $+ 1.0 \text{DISTS} + 0.1 \text{ICC} ]$} & \cmark & \cmark \\
      \midrule
      
      7 & 3 & 8 & $448 \times 256$ & 7 & \makecell{$5\cdot10^{-6}$\\$\downarrow$\\$10^{-6}$} & All & 0-15 & \makecell{16\\$\sim$\\384} & $\mathcal{L}_{RDPT}$ & \makecell[l]{$\mathbb{E}_t [ R + \lambda w_t(0.3 \text{MSE}$ \\ $+ 0.03 \text{LPIPS} + 0.03 \text{DISTS}$ \\ $+ 0.003 \text{ICC}) ]$} & \cmark & \cmark \\
      \bottomrule
  \end{tabular*}
\end{table}

\section{Test Details}\label{appendix:test}
This section provides comprehensive details of the experimental setup and evaluation protocols employed in this paper.

\paragraph{Performance Evaluation.} For compression performance, we conduct a quantitative analysis using rate-quality curves and report BD-Rate/BD-Metric values for both perceptual and distortion metrics. Qualitatively, we provide visualization results of the reconstructed video frames. Adhering to the input constraints of most NVC methods, we crop frame margins to ensure that both width and height are multiples of 64, following the protocol in~\cite{LSTVC,DiffVC,DiffVC-OSD}. Furthermore, given that the statistical properties of animation sequences in the MCL-JCV~\cite{MCL-JCV} dataset (Sequences 18, 20, 24, 25) diverge significantly from natural video content, we exclude these four sequences from our evaluation, consistent with the settings in~\cite{SSF,DCVC-TCM}. Unless otherwise stated, all evaluations are performed in the RGB color space.

\paragraph{Speed Evaluation.} To evaluate inference efficiency, we report encoding and decoding speeds (measured in frames per second, fps) across various resolutions and hardware devices. All reported speed metrics represent the average of multiple runs conducted on a strictly idle machine to ensure stability. Additionally, acknowledging that the inference latency of traditional codecs is highly sensitive to content complexity and quantization parameters (QP), we report their average throughput on the HEVC dataset averaged across four standard QPs (22, 27, 32, 37). It is worth emphasizing that all speed evaluations are conducted on data at the original resolution, without any cropping.

\paragraph{Configuration of Traditional Video Codecs.} For HM-16.25~\cite{HEVC} and VTM-17.0~\cite{VVC}, we utilize the standard \texttt{encoder\_lowdelay\_main\_rext.cfg} and \texttt{encoder\_lowdelay\_vtm.cfg} configuration files, respectively. Following the optimal testing pipeline demonstrated in DCVC-DC~\cite{DCVC-DC}, we convert RGB inputs to the YUV444 format as the intermediate processing color space for these codecs. We adopt the standard QP list of (22, 27, 32, 37). Detailed configuration arguments are provided below.
\begin{itemize}
  \item \textbf{HM}\\
  TAppEncoderStatic -c encoder\_lowdelay\_main\_rext.cfg 
  --InputFile=input\_path
  --BitstreamFile=bin\_path \\
  --DecodingRefreshType=2 
  --InputBitDepth=8 
  --OutputBitDepth=8 
  --OutputBitDepthC=8 \\
  --InputChromaFormat=444 
  --FrameRate=fps 
  --FramesToBeEncoded=96 
  --SourceWidth=width \\
  --SourceHeight=height 
  --IntraPeriod=32 
  --QP=qp 
  --Level=6.2
  \item \textbf{VTM}\\
  EncoderAppStatic -c encoder\_lowdelay\_vtm.cfg 
  --InputFile=input\_path 
  --BitstreamFile=bin\_path \\
  --DecodingRefreshType=2 
  --InputBitDepth=8 
  --OutputBitDepth=8 
  --OutputBitDepthC=8 \\
  --InputChromaFormat=444 
  --FrameRate=fps 
  --FramesToBeEncoded=96 
  --SourceWidth=width \\
  --SourceHeight=height 
  --IntraPeriod=32 
  --QP=qp 
  --Level=6.2
\end{itemize}

\paragraph{Configuration of NVC Methods.} We evaluate all NVC baselines using their officially released pre-trained models. For DCVC-FM~\cite{DCVC-FM} and DCVC-RT~\cite{DCVC-RT}, the quality scaling parameters (similar to QP) are set to (42, 49, 57, 63), maintaining YUV444 as the intermediate color space. For our proposed DiffVC-RT, the quality settings are configured as (0, 5, 10, 15).

\section{Supplementary Quantitative Results}\label{appendix:result}
\begin{figure*}[!t]
  \vskip 0.2in
  \centering
  \centerline{\includegraphics[width=\columnwidth]{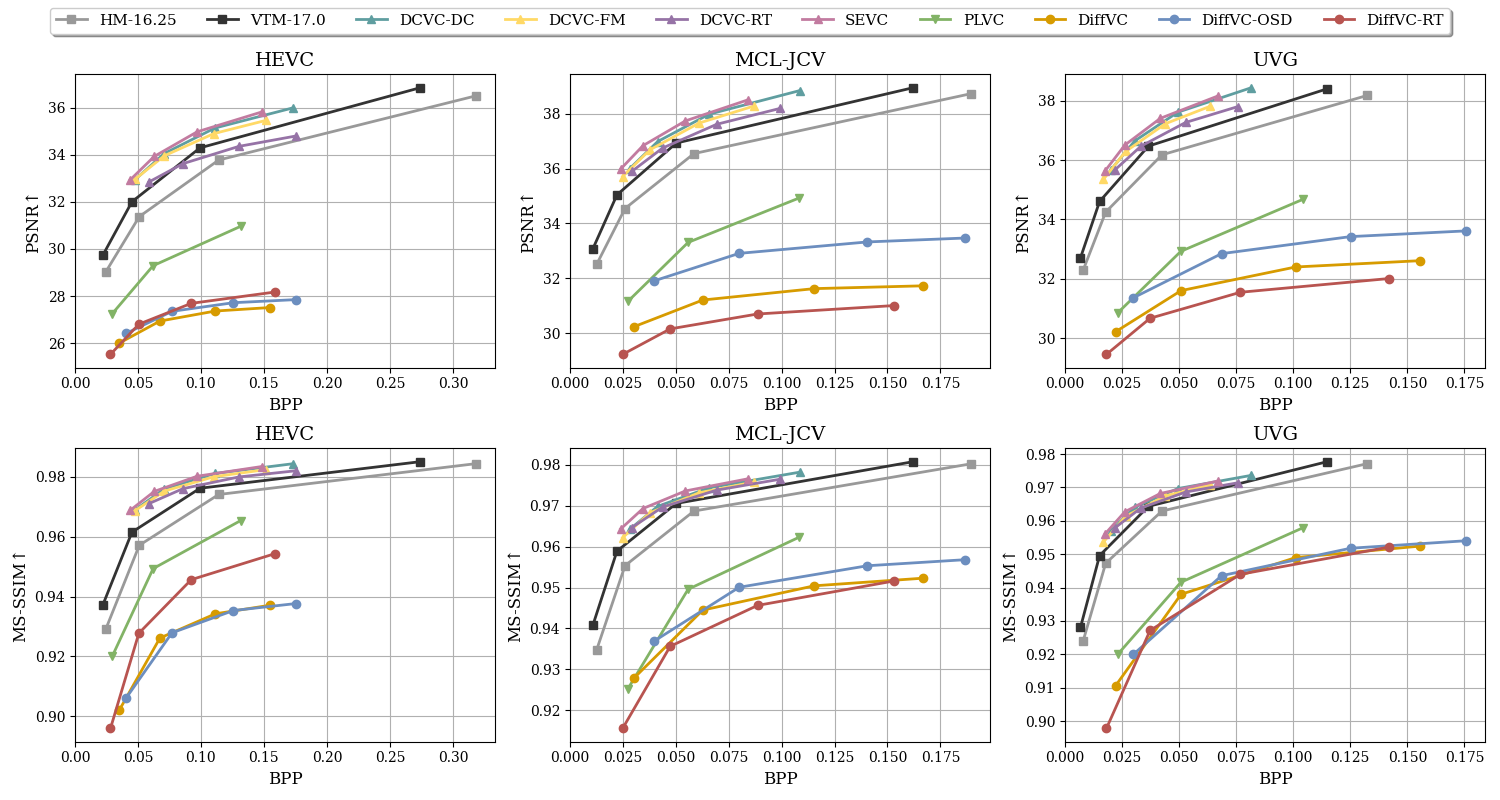}}
  \caption{
    Rate-distortion curves of the proposed DiffVC-RT and other methods on the HEVC, MCL-JCV, and UVG datasets.
  }
  \label{fig:additional_rd_curve}
\end{figure*}

This section provides a supplementary quantitative results of DiffVC-RT. Fig.~\ref{fig:additional_rd_curve} depicts the rate-distortion curves for DiffVC-RT and competing methods across three benchmark datasets. Consistent with established findings in the field of generative compression, diffusion-based methods generally trail behind traditional codecs and distortion-oriented NVCs on pixel-level metrics (e.g., PSNR and MS-SSIM~\cite{MS-SSIM}), as they prioritize perceptual quality over pixel-wise alignment. On the HEVC dataset~\cite{HEVC}, DiffVC-RT exhibits superior distortion performance compared to other diffusion-based methods. However, on high-resolution 1080p datasets such as MCL-JCV~\cite{MCL-JCV} and UVG~\cite{UVG}, its distortion performance is slightly lower than that of some diffusion-based baselines. This marginal degradation is an expected trade-off, attributable to the reduced model capacity of the pruned foundation model, which is explicitly used to unlock real-time inference capabilities.

Table~\ref{tab:results} details the specific BD-Rate and BD-Metric values for both perceptual and distortion metrics across all datasets, using VTM-17.0~\cite{VVC} as the anchor. Note that for all perceptual metrics where lower values denote better quality, we negate the scores during calculation. Consequently, in our tables, a lower (negative) BD-Rate and a higher (positive) BD-Metric indicate superior compression performance. In summary, DiffVC-RT consistently ranks among the top three across all perceptual metrics and benchmarks, successfully balancing competitive perceptual quality with significant breakthroughs in model complexity and inference efficiency.

Table~\ref{tab:complexity_detail} presents the parameter count, computational complexity, and latency of individual modules in DiffVC-RT. It is worth noting that we disabled the asynchronous and parallel strategies within the Asynchronous and Parallel Decoding Pipeline to accurately observe the intrinsic latency of each module. One can observe that the complexity of DiffVC-RT is primarily concentrated within the Frame Reconstructor, despite the utilization of a lightweight pruned diffusion model. Precisely for this reason, we propose the Asynchronous and Parallel Decoding Pipeline to mitigate the complexity imbalance between the Latent Compressor and the Frame Reconstructor, thereby significantly enhancing the overall inference throughput.

\begin{table*}[htbp]
	\centering
	\scriptsize
	\renewcommand {\arraystretch}{1.0}
	\caption{BD-Rate$\downarrow$ (\%) / BD-Metric$\uparrow$ for different methods on HEVC, MCL-JCV and UVG dataset. The anchor is VTM-17.0.}
	\label{tab:results}
	\resizebox{1.0\textwidth}{!}{
		\begin{threeparttable}
			\begin{tabularx}{\textwidth}{c >{\centering\arraybackslash}p{2.2cm} *{6}{>{\centering\arraybackslash}X}}
				\toprule
				\multicolumn{1}{c}{\multirow{2}{*}[-0.5ex]{\textbf{Dataset}}} & 
				\multicolumn{1}{c}{\multirow{2}{*}[-0.5ex]{\textbf{Method}}} & 
				\multicolumn{3}{c}{\textbf{Perception}} & 
				\multicolumn{2}{c}{\textbf{Distortion}}  \\
				\cmidrule(lr){3-5} \cmidrule(lr){6-7}
				& & \textbf{LPIPS} & \textbf{DISTS} & \textbf{FloLPIPS} & \textbf{PSNR} & \textbf{MS-SSIM} \\
				\midrule
				\multirow{10}{*}{HEVC}
        & HM-16.25 & 17.6 / -0.0088 & 0.7 / -0.0002 & 18.9 / -0.0088 & 37.8 / -0.9235 & 33.0 / -0.0057 \\
        & VTM-17.0 & 0.0 / 0.0000 & 0.0 / 0.0000 & 0.0 / 0.0000 & 0.0 / 0.0000 & 0.0 / 0.0000 \\
        & DCVC-DC & -7.0 / 0.0029 & 25.2 / -0.0073 & 2.1 / -0.0011 & \textbf{\textcolor{myblue}{-19.8 / 0.5900}} & \textbf{\textcolor{myblue}{-28.5 / 0.0042}} \\
        & DCVC-FM & -7.3 / 0.0032 & 21.0 / -0.0063 & 2.9 / -0.0014 & \textbf{\textcolor{mygreen}{-16.4 / 0.4788}} & \textbf{\textcolor{mygreen}{-23.3 / 0.0036}} \\
        & DCVC-RT & 10.0 / -0.0042 & 36.1 / -0.0097 & 20.5 / -0.0077 & 15.7 / -0.4340 & -12.1 / 0.0014 \\
        & SEVC & -4.9 / 0.0022 & 25.7 / -0.0078 & 13.6 / -0.0058 & \textbf{\textcolor{myred}{-25.6 / 0.8075}} & \textbf{\textcolor{myred}{-32.0 / 0.0052}} \\
        & PLVC & \textbf{\textcolor{mygreen}{-66.9 / 0.0518}} & -27.1 / 0.0110 & -79.7 / 0.0725 & 262.4 / -3.7475 & 120.0 / -0.0209 \\
        & DiffVC & -64.8 / 0.0401 & \textbf{\textcolor{myblue}{-79.0 / 0.0391}} & \textbf{\textcolor{mygreen}{-83.3 / 0.0695}} & N/A / -6.4813 & N/A / -0.0453 \\
        & DiffVC-OSD & \textbf{\textcolor{myblue}{-77.1 / 0.0457}} & \textbf{\textcolor{myred}{N/A / 0.0428}} & \textbf{\textcolor{myblue}{N/A / 0.0781}} & N/A / -6.4733 & 657.8 / -0.0457 \\
        & DiffVC-RT & \textbf{\textcolor{myred}{-80.1 / 0.0600}} & \textbf{\textcolor{mygreen}{-68.7 / 0.0314}} & \textbf{\textcolor{myred}{N/A / 0.0892}} & N/A / -6.0052 & 258.0 / -0.0347 \\ \midrule
        \multirow{10}{*}{MCL-JCV}
        & HM-16.25 & 23.7 / -0.0118 & 0.6 / -0.0001 & 31.0 / -0.0141 & 40.2 / -0.7545 & 37.5 / -0.0049 \\
        & VTM-17.0 & 0.0 / 0.0000 & 0.0 / 0.0000 & 0.0 / 0.0000 & 0.0 / 0.0000 & 0.0 / 0.0000 \\
        & DCVC-DC & 56.6 / -0.0213 & 95.8 / -0.0208 & 86.5 / -0.0281 & \textbf{\textcolor{myblue}{-22.3 / 0.4831}} & \textbf{\textcolor{myblue}{-10.4 / 0.0011}} \\
        & DCVC-FM & 58.5 / -0.0224 & 90.2 / -0.0210 & 94.1 / -0.0309 & \textbf{\textcolor{mygreen}{-16.1 / 0.3549}} & \textbf{\textcolor{mygreen}{-7.0 / 0.0008}} \\
        & DCVC-RT & 73.9 / -0.0263 & 103.4 / -0.0221 & 99.3 / -0.0314 & -4.4 / 0.0851 & -3.5 / 0.0003 \\
        & SEVC & 56.4 / -0.0217 & 89.6 / -0.0211 & 87.7 / -0.0293 & \textbf{\textcolor{myred}{-28.6 / 0.6769}} & \textbf{\textcolor{myred}{-21.0 / 0.0026}} \\
        & PLVC & N/A / 0.0857 & -67.6 / 0.0300 & N/A / 0.0930 & 384.6 / -3.9255 & 296.0 / -0.0247 \\
        & DiffVC & \textbf{\textcolor{myblue}{N/A / 0.0868}} & \textbf{\textcolor{myblue}{N/A / 0.0556}} & \textbf{\textcolor{mygreen}{N/A / 0.1028}} & N/A / -6.3123 & 509.2 / -0.0296 \\
        & DiffVC-OSD & \textbf{\textcolor{myred}{N/A / 0.1015}} & \textbf{\textcolor{myred}{N/A / 0.0694}} & \textbf{\textcolor{myred}{N/A / 0.1231}} & 1026.2 / -4.9601 & 444.4 / -0.0261 \\
        & DiffVC-RT & \textbf{\textcolor{mygreen}{N/A / 0.0858}} & \textbf{\textcolor{mygreen}{-80.1 / 0.0361}} & \textbf{\textcolor{myblue}{N/A / 0.1074}} & N/A / -6.9416 & 642.1 / -0.0339 \\ \midrule
        \multirow{10}{*}{UVG}
        & HM-16.25 & 3.6 / -0.0014 & -16.1 / 0.0048 & 7.6 / -0.0042 & 36.0 / -0.6329 & 29.2 / -0.0046 \\
        & VTM-17.0 & 0.0 / 0.0000 & 0.0 / 0.0000 & 0.0 / 0.0000 & 0.0 / 0.0000 & 0.0 / 0.0000 \\
        & DCVC-DC & 43.5 / -0.0168 & 89.4 / -0.0160 & 27.2 / -0.0156 & \textbf{\textcolor{myblue}{-25.8 / 0.5460}} & \textbf{\textcolor{myblue}{-11.6 / 0.0015}} \\
        & DCVC-FM & 38.1 / -0.0151 & 95.1 / -0.0181 & 41.9 / -0.0210 & \textbf{\textcolor{mygreen}{-20.4 / 0.4366}} & \textbf{\textcolor{mygreen}{-8.1 / 0.0011}} \\
        & DCVC-RT & 49.9 / -0.0185 & 102.4 / -0.0177 & 44.5 / -0.0222 & -8.1 / 0.1521 & -2.0 / 0.0000 \\
        & SEVC & 43.4 / -0.0168 & 105.8 / -0.0194 & 43.7 / -0.0217 & \textbf{\textcolor{myred}{-30.6 / 0.6825}} & \textbf{\textcolor{myred}{-16.3 / 0.0023}} \\
        & PLVC & \textbf{\textcolor{myblue}{N/A / 0.0873}} & N/A / 0.0336 & \textbf{\textcolor{mygreen}{N/A / 0.0994}} & 568.5 / -4.2150 & 374.9 / -0.0284 \\
        & DiffVC & -78.6 / 0.0722 & \textbf{\textcolor{myblue}{N/A / 0.0464}} & -77.0 / 0.0873 & N/A / -5.5030 & 486.3 / -0.0330 \\
        & DiffVC-OSD & \textbf{\textcolor{myred}{N/A / 0.0913}} & \textbf{\textcolor{myred}{N/A / 0.0621}} & \textbf{\textcolor{myred}{N/A / 0.1141}} & 1060.3 / -4.7638 & 498.6 / -0.0316 \\
        & DiffVC-RT & \textbf{\textcolor{mygreen}{N/A / 0.0788}} & \textbf{\textcolor{mygreen}{-79.9 / 0.0356}} & \textbf{\textcolor{myblue}{N/A / 0.1049}} & N/A / -5.9737 & 540.7 / -0.0369 \\ \bottomrule
			\end{tabularx}
			\begin{tablenotes}
				\item[*] \textbf{\textcolor{myred}{Red}}, \textbf{\textcolor{myblue}{Blue}} and \textbf{\textcolor{mygreen}{Green}} indicate the best, second-best and third-best performance, respectively. 'N/A' indicates that BD-rate cannot be calculated due to the lack of overlap. 
			\end{tablenotes}
		\end{threeparttable}
	}
\end{table*}

\begin{table*}[t]
  \caption{
    Detailed complexity analysis of DiffVC-RT. The reported Time (ms) denotes the average per-inter-frame runtime measured on 480p video sequences using a single NVIDIA H800 GPU.
  }
  \label{tab:complexity_detail}
  \centering
  \scriptsize   
  \setlength{\tabcolsep}{4pt}
  \renewcommand{\arraystretch}{1.05}

  \begin{tabular*}{0.75\columnwidth}{@{\extracolsep{\fill}}lcccc@{}}
  \toprule
  \multirow{2}{*}{\textbf{DiffVC-RT}} & \multirow{2}{*}{\textbf{Latent Compressor}} & \multicolumn{2}{c}{\textbf{Frame Reconstructor}} & \multirow{2}{*}{\textbf{All}} \\
  \cmidrule(lr){3-4} 
  & & \textbf{U-Net} & \textbf{VAE Decoder} & \\
  \midrule
  Params (M) & 17.10 & 444.78 & 12.38 & 474.26 \\
  kMACs/pixel & 141.32 & 1072.08 & 1248.96 & 2462.36 \\
  Time (ms) & 4.08 & 19.32 & 20.66 & 44.06 \\ \bottomrule
  \end{tabular*}
\end{table*}

\section{Supplementary Ablation Studies}\label{appendix:ablation}
\subsection{Efficient \& Informative Model Architecture (EIMA)}
\begin{figure*}[!t]
  \vskip 0.2in
  \centering
  \centerline{\includegraphics[width=\columnwidth]{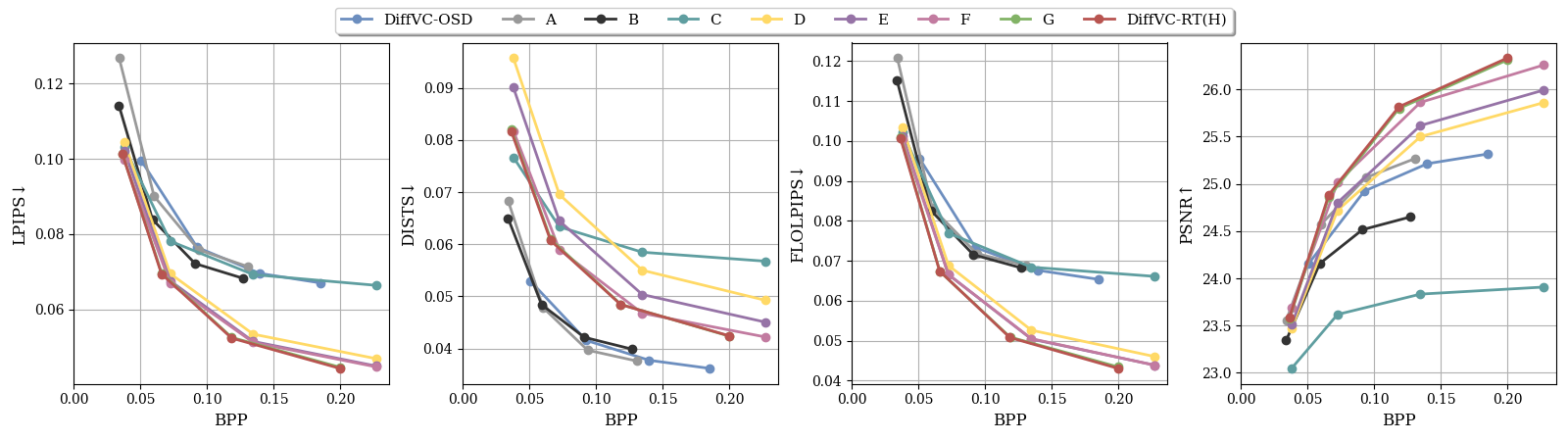}}
  \caption{
    Rate-quality curves of different ablation variants on HEVC Class C.
  }
  \label{fig:ablation_rd_curve}
\end{figure*}

Fig.~\ref{fig:ablation_rd_curve} presents the rate-quality curves for various ablation variants on the HEVC Class C dataset, with variant naming consistent with Table~\ref{tab:ablation}. By synthesizing the quantitative results from Table~\ref{tab:ablation} and the visual trends in Fig.~\ref{fig:ablation_rd_curve}, we provide a detailed analysis of the proposed Efficient and Informative Model Architecture:

\paragraph{DiffVC-OSD vs. Model A.} Despite eliminating the motion branch present in the baseline DiffVC-OSD~\cite{DiffVC-OSD}, Model A achieves compression performance comparable to the baseline. This finding suggests that the motion branch contributes negligibly to the efficacy of diffusion-based NVC frameworks.

\paragraph{Model A vs. Model B.} Incorporating the lightweight Pruned U-Net and VAE Decoder enables Model B to surpass Model A in perceptual metrics. However, this efficiency gain comes at the cost of distortion performance, which degrades due to the reduced model capacity.

\paragraph{Model B vs. Model C.} Removing the pre-trained VAE Encoder in Model C mitigates structural information loss at the encoding side. However, a narrow channel bottleneck persists between the Latent Compressor and the Frame Reconstructor, as well as between the U-Net and VAE Decoder. Consequently, Model C exhibits an exceptionally flat rate-quality curve with a distinct performance ceiling, a trend particularly evident in the rate-PSNR curve. Notably, variants derived from Model C yield inferior performance compared to Models A/B and DiffVC-OSD~\cite{DiffVC-OSD} on the DISTS~\cite{DISTS} metric. This degradation arises because the absence of the pre-trained VAE Encoder diminishes the generative priors typically provided by such powerful feature extractors. Nevertheless, we accept this trade-off as a necessary compromise to unlock real-time inference capabilities.

\paragraph{Model C vs. Model D.} As depicted in Fig.~\ref{fig:ablation_rd_curve}, Model D demonstrates a significantly steeper rate-quality curve and a higher performance ceiling compared to Models A/B/C and the baseline. This substantial improvement is attributed to the Latent Channel Expansion strategy, which effectively establishes a high-capacity end-to-end information pathway.

\begin{figure*}[t]
  \vskip 0.2in
  \centering
  
  \begin{subfigure}[b]{0.45\linewidth}
    \centering
    \includegraphics[width=\linewidth]{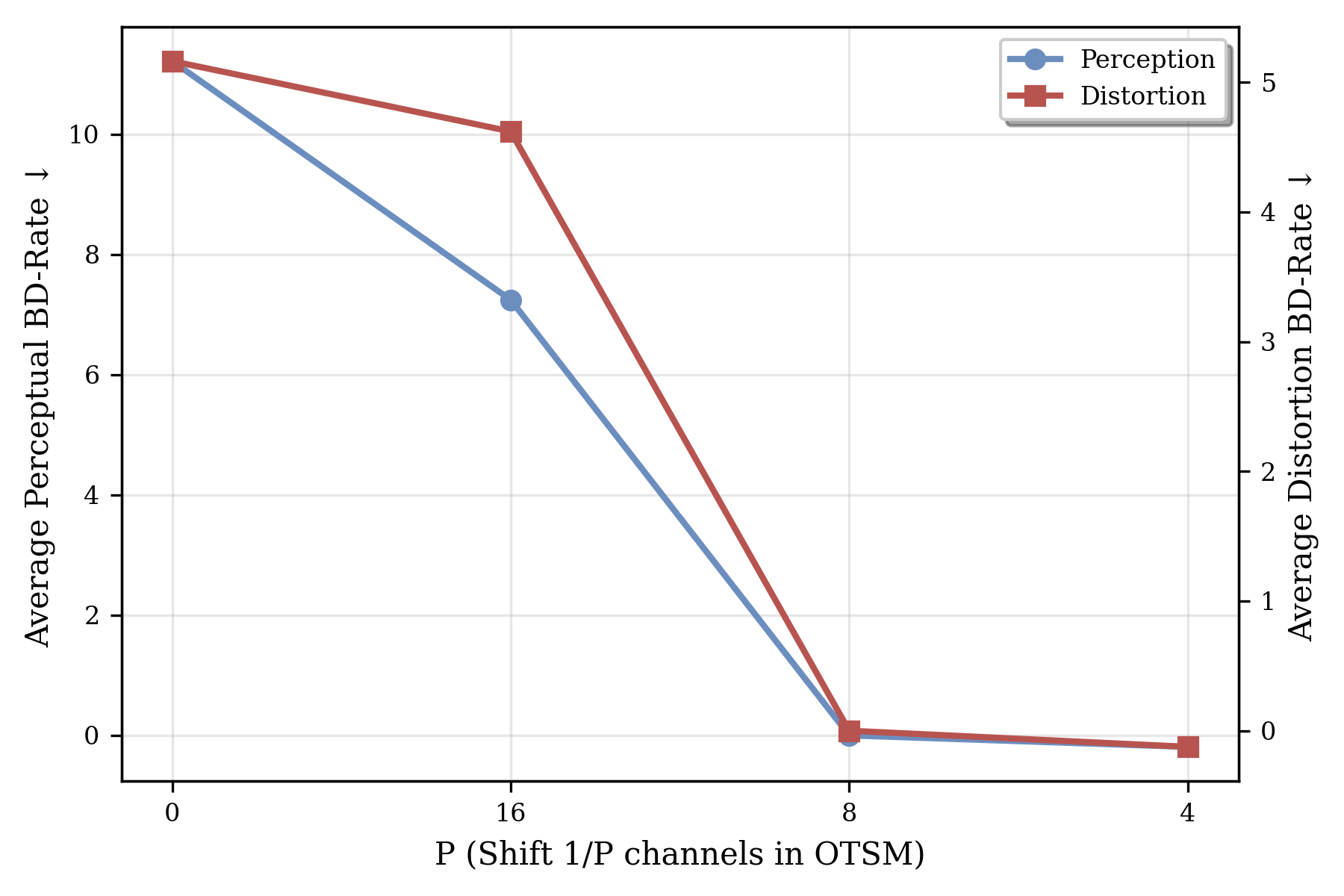} 
    \caption{} 
    \label{fig:eicm_ablation_curve_a}
  \end{subfigure}
  \hfill
  \begin{subfigure}[b]{0.45\linewidth}
    \centering
    \includegraphics[width=\linewidth]{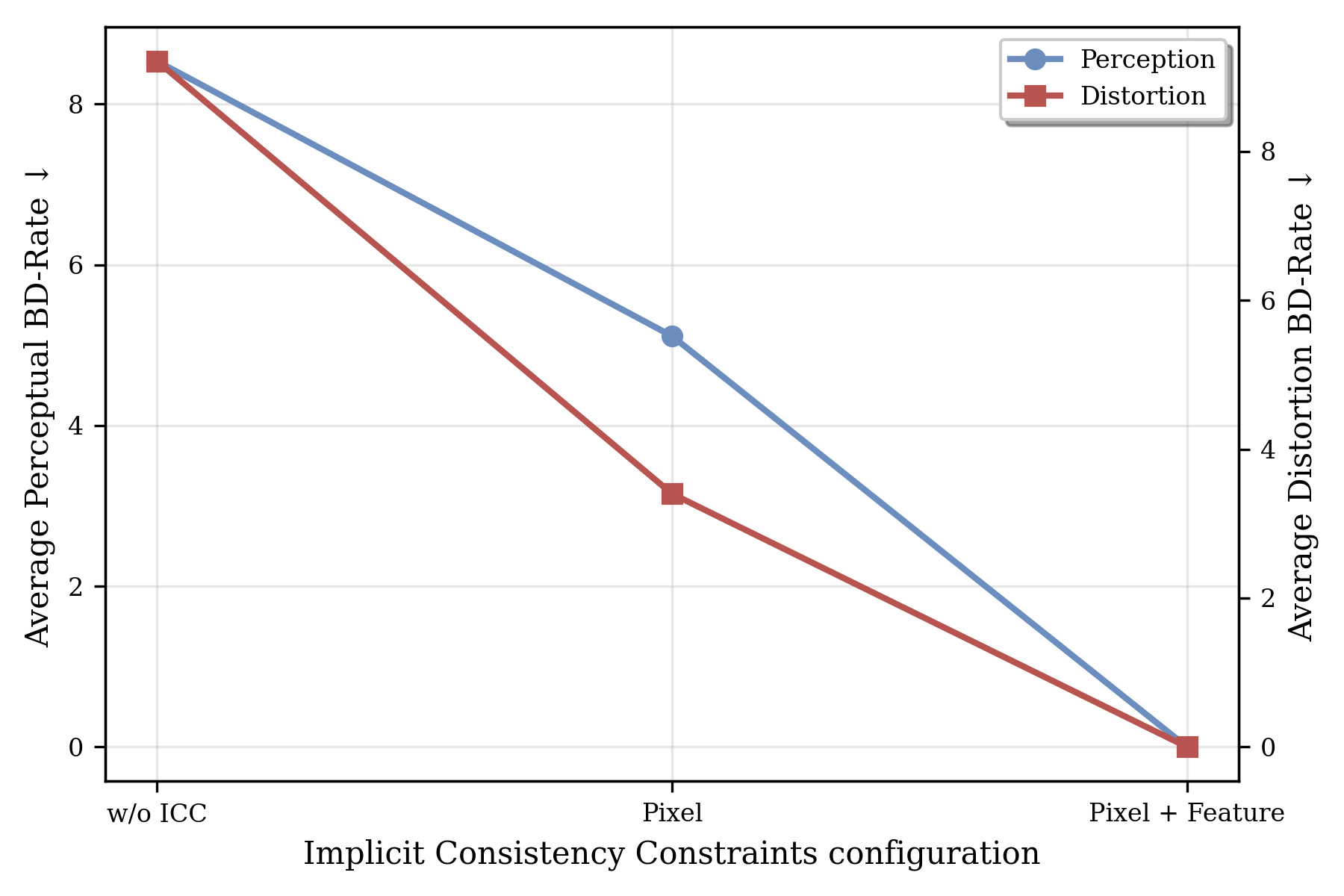}
    \caption{}
    \label{fig:eicm_ablation_curve_b}
  \end{subfigure}
  
  \caption{
    Ablation studies on the Explicit and Implicit Consistency Modeling.
    (a) Impact of the channel shift proportion ($1/P$) in the Online Temporal Shift Module (Explicit Consistency Enhancement) on performance.
    (b) Performance comparison of different constraint types employed in the Implicit Consistency Constraints.
  }
  \label{fig:eicm_ablation_curve}
\end{figure*}

\subsection{Explicit \& Implicit Consistency Modeling (EICM)}
We conduct a more comprehensive ablation study regarding Explicit and Implicit Consistency Modeling. Fig.~\ref{fig:eicm_ablation_curve} illustrates the performance of Explicit Consistency Enhancement and Implicit Consistency Constraints under various configurations.

As depicted in Fig.~\ref{fig:eicm_ablation_curve_a}, increasing the channel shift proportion ($1/P$) strengthens the intensity of inter-frame interactions. Consequently, performance in terms of both perceptual and distortion metrics shows a steady improvement before gradually stabilizing. Based on these observations, we select $P=8$ as the default configuration for DiffVC-RT.

Fig.~\ref{fig:eicm_ablation_curve_b} demonstrates that temporal consistency constraints in both pixel and feature spaces positively contribute to perceptual and distortion performance. Furthermore, their combination yields a synergistic effect, resulting in optimal performance. Accordingly, DiffVC-RT adopts a hybrid strategy that integrates both constraints.

\subsection{Asynchronous \& Parallel Decoding Pipeline (APDP)}
We conducted a detailed ablation study on the proposed Asynchronous and Parallel Decoding Pipeline (APDP). Table~\ref{tab:apdp_ablation} presents the decoding speeds of DiffVC-RT under various resolutions, devices, and APDP configurations. The results show that decoupling the Latent Compressor and Frame Reconstructor via asynchronous execution yields a tangible increase in throughput. Moreover, employing Parallel Frame Reconstruction provides further acceleration to the decoding process.

However, decoding speed saturates as the parallel frame number $N$ increases, suggesting a \textbf{bottleneck shift}: from kernel-launch overhead and memory latency to saturated GPU compute. As the parallel workload grows, Tensor Core utilization nears its peak, limiting further throughput scaling. We set $N=8$ on NVIDIA H800/A800 and $N=4$ on the RTX 3090 to balance decoding latency and throughput. APDP remains configurable for specific hardware and application constraints.

\begin{table*}[t]
  \caption{
    Ablation study on the Asynchronous and Parallel Decoding Pipeline. We report the decoding speed (fps) of DiffVC-RT under different APDP configurations across various devices and resolutions. “OOM” indicates out-of-memory conditions.
  }
  \label{tab:apdp_ablation}
  \centering
  \scriptsize   
  \setlength{\tabcolsep}{4pt}
  \renewcommand{\arraystretch}{1.05}

  \begin{tabular*}{\columnwidth}{@{\extracolsep{\fill}}lccccccc@{}}
  \toprule
  \multirow{3}{*}{\textbf{Device}} & \multirow{3}{*}{\textbf{Resolution}} & \multicolumn{6}{c}{\textbf{APDP configuration}} \\ \cmidrule(lr){3-8}
  & & \multirow{2}{*}{\textbf{\makecell{w/o Asynchronization\\+ w/o Parallelization}}} & \textbf{w/ Asynchronization + w/o Parallelization} & \multicolumn{4}{c}{\textbf{w/ Asynchronization + w/ Parallelization}} \\ \cmidrule(lr){4-4} \cmidrule(lr){5-8} 
  & & & \textbf{N=1} & \textbf{N=2} & \textbf{N=4} & \textbf{N=8} & \textbf{N=16} \\
  \midrule
  \multirow{2}{*}{NVIDIA H800} & 480p & 23.75 & 39.53 & 54.32 & 61.30 & 74.79 & 75.06 \\
  & 1080p & 5.65 & 9.48 & 10.30 & 10.37 & 10.95 & 10.70 \\
  \midrule
  \multirow{2}{*}{RTX 3090} & 480p & 13.77 & 17.26 & 19.28 & 21.11 & 24.35 & 24.17  \\
  & 1080p & 2.53 & 3.00 & 3.10 & 3.13 & OOM & OOM \\
  \bottomrule
  \end{tabular*}
\end{table*}

\begin{figure*}[!t]
  \vskip 0.2in
  \centering
  \centerline{\includegraphics[width=\columnwidth]{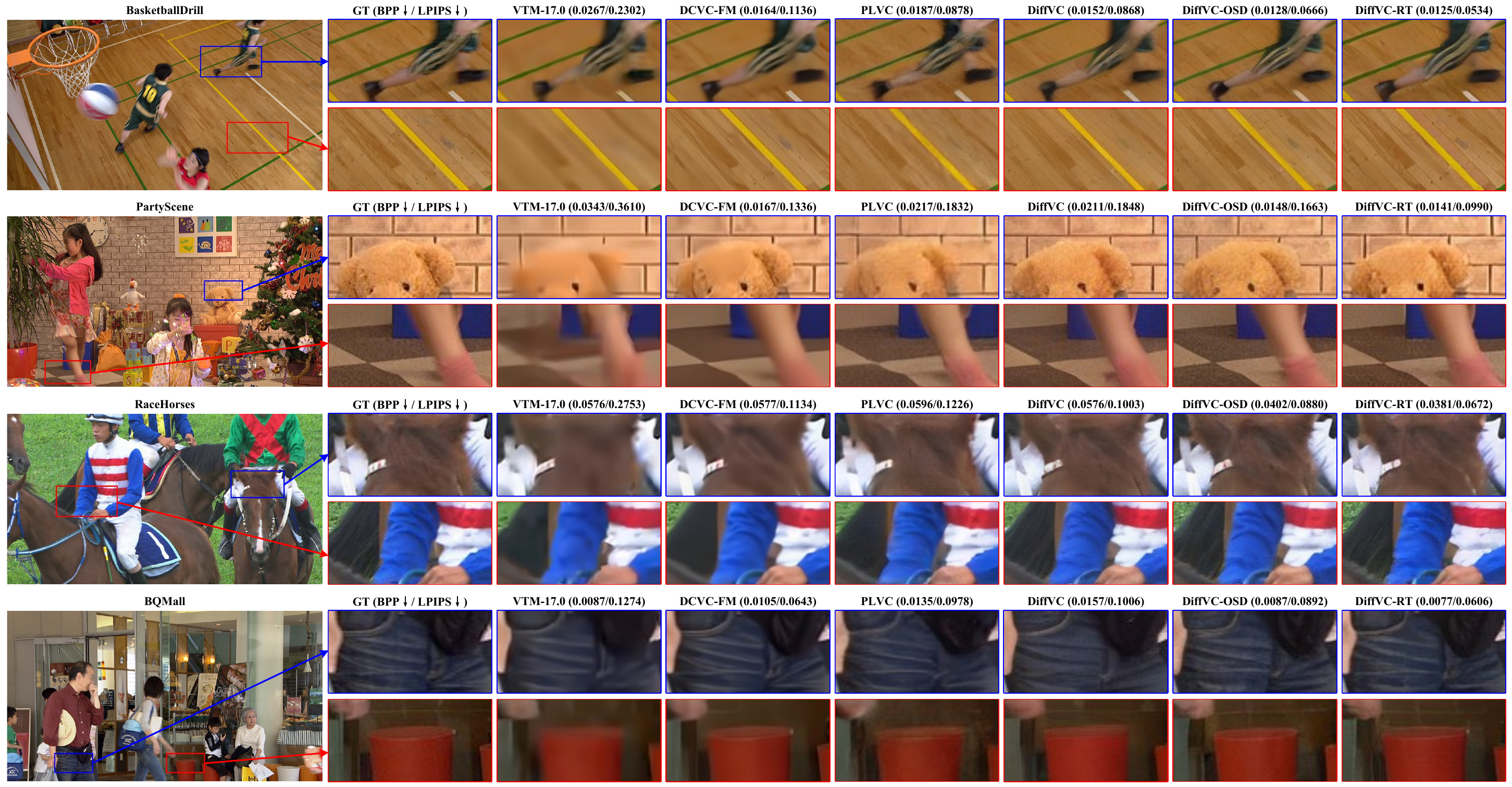}}
  \caption{
    Additional visual comparisons between the proposed DiffVC-RT and other methods.
  }
  \label{fig:additional_visual_result}
\end{figure*}

\section{Additional Visual Examples}\label{appendix:visual}
Fig.~\ref{fig:additional_visual_result} presents additional visual comparisons between DiffVC-RT and various other methods, focusing on reconstruction details across diverse scenes. Consistent with previous conclusions, DiffVC-RT reconstructs video frames with richer details and superior clarity, even at lower bitrates.


\end{document}